%% file: main.tex
\documentclass[conference]{IEEEtran}
\usepackage[utf8]{inputenc}
\usepackage[german,american]{babel}
\usepackage[T1]{fontenc}
\usepackage{cite}
\usepackage{amsmath,amssymb,amsfonts}
\usepackage{algorithmic}
\usepackage{graphicx}
\usepackage{textcomp} 
\usepackage{xcolor}
\usepackage{tabularx}
\usepackage{array}
\usepackage{booktabs}
\usepackage{enumitem}
\usepackage{mathrsfs}
\usepackage{wasysym}
\usepackage{makecell} 
\usepackage[bookmarks=false]{hyperref}

\begin{document}
\bstctlcite{IEEEexample:BSTcontrol} 

\title{A Research Agenda for AI Planning in the Field of Flexible Production Systems}

\author{
\IEEEauthorblockN{
    Aljosha Köcher\IEEEauthorrefmark{1},
    René Heesch\IEEEauthorrefmark{1},
    Niklas Widulle\IEEEauthorrefmark{1},
    Anna Nordhausen\IEEEauthorrefmark{1},\\
    Julian Putzke\IEEEauthorrefmark{1},
    Alexander Windmann\IEEEauthorrefmark{1},
    Oliver Niggemann\IEEEauthorrefmark{1}
}
\IEEEauthorblockA{
\IEEEauthorrefmark{1}Institute of Automation\\
Helmut Schmidt University, Hamburg, Germany\\
Email: firstName.lastName@hsu-hh.de}
}

\maketitle

\begin{abstract}
Manufacturing companies face challenges when it comes to quickly adapting their production control to fluctuating demands or changing requirements. Control approaches that encapsulate production functions as services have shown to be promising in order to increase the flexibility of Cyber-Physical Production Systems. But an existing challenge of such approaches is finding a production plan based on provided functionalities for a demanded product, especially when there is no direct (i.e., syntactic) match between demanded and provided functions.
While there is a variety of approaches to production planning, flexible production poses specific requirements that are not covered by existing research. 
In this contribution, we first capture these requirements for flexible production environments. Afterwards, an overview of current Artificial Intelligence approaches that can be utilized in order to overcome the aforementioned challenges is given.
For this purpose, we focus on planning algorithms, but also consider models of production systems that can act as inputs to these algorithms. Approaches from both symbolic AI planning as well as approaches based on Machine Learning are discussed and eventually compared against the requirements. Based on this comparison, a research agenda is derived.
\end{abstract}


\input{intro.tex}

\input{reqs.tex}

\input{models.tex}

\input{symbolic.tex}

\input{ml.tex}

\input{scheduling.tex}

\input{discussion.tex}

\input{conclusion.tex}

\bibliographystyle{IEEEtran}
\bibliography{references} 

\end{document}

%% file: intro.tex
\section{Introduction}
\label{sec:introduction}

Today's companies operate in an environment characterized by more frequently changing customer demands leading to shorter product life cycles and smaller batch sizes. In order to perform successfully in this market, companies have to adjust their production quickly to changing requirements. This is often referred to as \emph{changeability} or \emph{adaptability} of production systems \cite{wiendahl2007changeable}. Changeability covers both changes on a physical as well as on control level. While the first contains ways to add or remove machines and components of a plant, the second covers adaptations in process planning or production control \cite{ElM_ReconfigurableProcessPlansFor_2007}. This paper deals with the second aspect and focuses on planning of process sequences on the shop-floor.

For this purpose, one class of highly changeable production systems are so-called \emph{Cyber-Physical Production Systems} (CPPS). CPPS with their autonomous and cooperative behavior regarding all levels of production break with traditional, more hierarchical automation approaches \cite{Mon_CyberphysicalProductionSystems:Roots_2014}.
But in order to exploit the potential of an increased changeability on control level, production planning needs to be automated for use in CPPS. AI planning approaches, which are concerned with finding a sequence of actions that lead from an initial to a goal state \cite{RuNo_Artificialintelligence_2016}, have long been researched for production planning, but current planning systems do not satisfy the requirements of CPPS \cite{rogalla_improved_2018}.

\begin{figure*}[htb]
    \label{fig:planning-steps} 
    \centering
    \includegraphics[width=\textwidth]{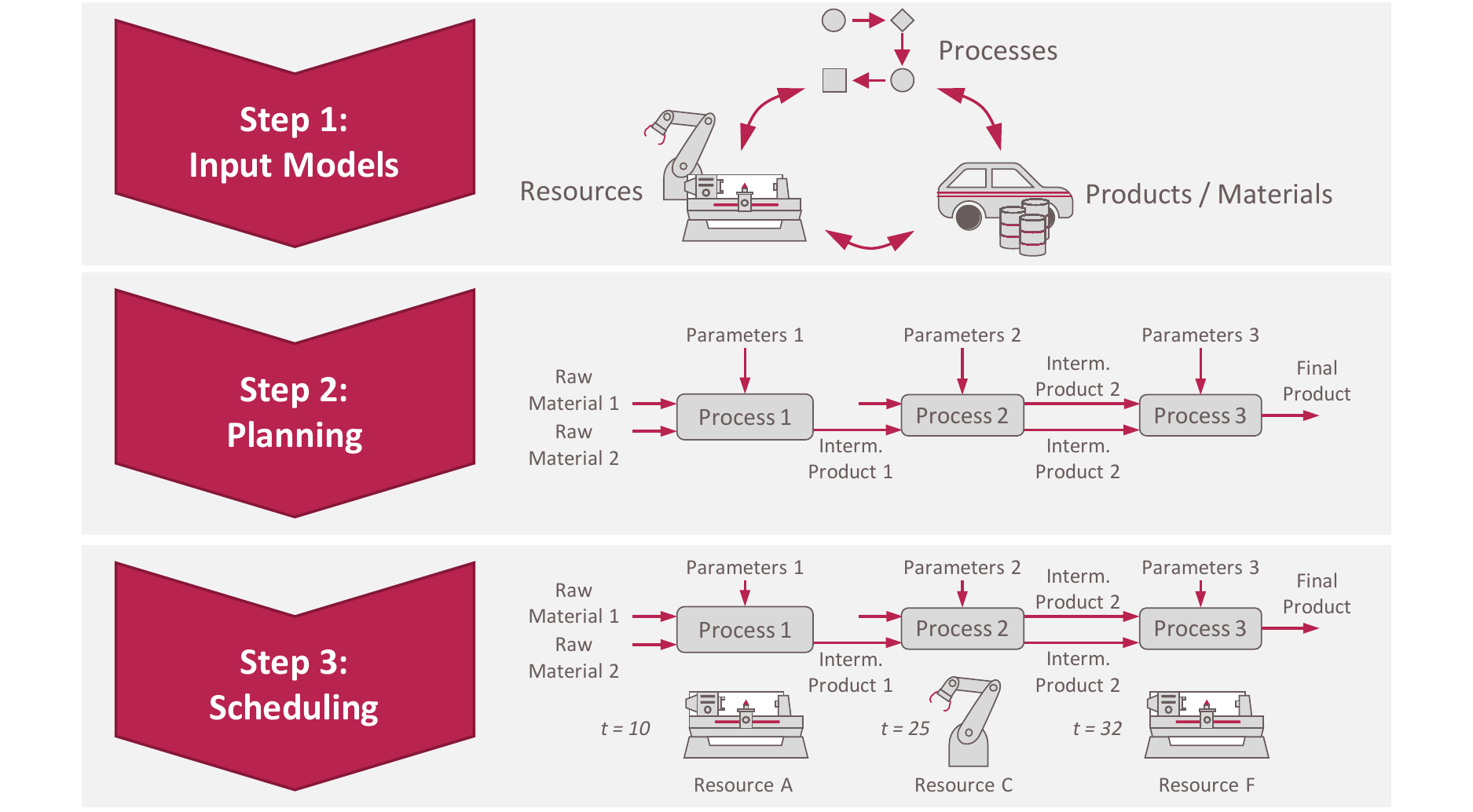}
    \caption{Typical steps to obtain an executable production plan. After modeling a system in a preliminary first step, a sequence of action needs to be found (Step 2) before resource allocation and scheduling (Step 3) is done.}
\end{figure*}
\smallskip

An ideal AI planning solution to find an executable production plan for a given order can be divided into the three main steps of Fig. \ref{fig:planning-steps}.
A first---and typically preliminary---step is concerned with creating a model of the production system (see Step 1 in Fig. \ref{fig:planning-steps}). Such models are often created with regards to the three views \emph{product}, \emph{process} and \emph{resource} \cite{ScDr_Threeviewconceptformodelingprocess_2009}. Production resources, their components and connections to other resources are modeled to capture dependencies on a structural level. Process steps are defined by their inputs (e.g. products, energy, information), by their outputs and by parameters. Parameters such as timings and speeds configure the used resources. Product descriptions include raw materials, intermediate products and especially the final product.
Although there is still a heterogeneous variety of tools and resulting models, semantic models based on ontologies are increasingly getting used to represent the complex interrelationships in the context of CPPS. 
For planning and scheduling, an additional part of input models are optimization criteria, which could be CO\textsubscript{2} generation, resource consumption, price or throughput. Both the actual model as well as optimization criteria act as inputs for subsequent AI algorithms. Based on these inputs and for a given product to be manufactured, a valid process sequence should be computed automatically by an AI system.

Once an input model has been created, AI algorithms for such problems first solve a planning problem (see Step 2 in Fig. \ref{fig:planning-steps}): All necessary process steps leading from initial input products to the final product are determined. This includes computing the features of all intermediate products such as dimensions and the computation of optimal parameters for the steps, e.g. timings or transport speeds.

As a third and final step, a scheduling problem is solved in order to find an optimal process sequence with respect to the optimization criteria given (see Step 3 in Fig. \ref{fig:planning-steps}): Each process step is mapped to a resource and start and end times are computed for all process steps on every machine.

\medskip
Currently, the modeling tasks and the planning/scheduling steps are typically done manually by experts---leading to high efforts, long downtimes and potentially sub-optimal results. Furthermore, CPPS models are seldom created with subsequent planning algorithms in mind. 
And vice versa, current planning algorithms do not accept semantic CPPS models of current research as input. This leads to the main research question: \emph{Can existing AI solutions on all three steps be interconnected and used to achieve an automated AI planning solutions for CPPS?} 

In \cite{ESA+_RoadmaptoSkillBased_2019}, the authors describe initial ideas to use a classical planning approach in a flexible production. We would like evaluate a broader set of approaches for their applicability in the domain of CPPS.
Currently, the two areas of CPPS modeling and planning are rather disjoint with a focus on rigorous representation on one side and a focus on algorithms on the other side. One contribution of this paper is to establish a common notion between these two areas in which modeling and planning are understood as interrelated steps of an AI solution to production planning (compare Fig. \ref{fig:planning-steps}).
The second contribution is an assessment of current research with regard to these solution steps and an identification of \emph{research gaps}.

The remainder of this paper is structured as follows. In Section \ref{sec:requirements}, requirements to AI solutions for CPPS are presented. Section \ref{sec:ai} summarizes existing AI approaches to the three steps described above: 
While Section \ref{sec:models} covers related work in the field of CPPS modeling, the subsequent subsections contain an overview of AI algorithms for the two algorithmic steps of planning and scheduling. Approaches to all three steps are compared with the requirements and discussed in Section \ref{sec:discussion}. Afterwards, the derived research agenda is presented before a short summary concludes this paper.

%% file: reqs.tex
\section{Requirements}
\label{sec:requirements}
We identified six requirements that an AI-approach covering all steps shown in Fig. \ref{fig:planning-steps}  needs to meet. 

\textbf{(R1)} Every step must support optimization with regard to different criteria and even contradicting cost functions such as resource consumption, CO\textsubscript{2} generation, price or throughput. 

\textbf{(R2)} Complex dependencies between inputs, outputs and process parameters must be taken into account. Such dependencies might be expressed as a symbolic relation between terms but could also be expressed as functional relations. One example is a chemical process creating varying product properties (e.g. viscosity) depending on the process parameters (e.g. temperature over time).
 
\textbf{(R3)} Recurring calls of functionalities such as loops in the control flow need to be considered. For example in a painting process, it might be necessary to paint an area more than once to reach a certain layer thickness. Even better than detecting multiple calls would be to explicitly detect loops including the underlying conditions, as this could prevent replanning if demanded properties of a product change.

\textbf{(R4)} The fourth requirement we identified concerns the \emph{explainability} of an AI planning solution. Explainability includes the transparency of a planner's decisions as well as the trust in its functionality. It is necessary that production workers and managers trust a novel planning approach in the same way they trust current production planning systems. Thus the decisions of a planning solution have to be explainable and planning errors have to be recognizable and correctable \cite{Beyerer.2021}.

\textbf{(R5)} Implementing a planning solution requires knowledge and effort by an automation engineer. The amount of implementation effort required is dependent on the chosen solution. Therefore a further requirement is to keep this effort low to maintain economic viability and to gain a net benefit from introducing an AI approach.

\textbf{(R6)} There is a variety of various information and data coming from existing systems such as Enterprise Resource Planning or Manufacturing Execution Systems (MES). This data is challenging to work with since real world data is often noisy, incomplete and has defects. Existing data needs to be taken into account and should be used to automate the planning process in all steps.

%% file: models.tex
\section{Existing AI Approaches}
\label{sec:ai}

\subsection{Semantic Models} \label{sec:models}

CPPS need to interoperate with each other and with other systems. Semantic Web standards and especially ontologies are considered to be solutions with a high potential to represent knowledge about CPPS in a machine-interpretable way. \cite{FGL+_SemanticCPPSinIndustry_2020}

Models of symbolic AI such as ontologies allow for an abstract modeling of entities, their properties and the relations between entities. 
Furthermore, they support advanced querying and reasoning, i.e. automatic computation of so-far unknown facts.
For production planning and control, research approaches from the field of so-called \emph{capability- and skill-based production} are seen as a promising means to model CPPS and their functionalities. These approaches use semantic models for formal descriptions of resources, their capabilities and skills. While capabilities are understood as abstract process descriptions in these approaches, skills are seen as executable machine functions, which can be invoked, e.g., via OPC UA. 
A rather generic meta-model focusing on the capability aspect is presented in \cite{WBS+_AnOntologybasedMetamodelfor_2020} as an ontology that contains only very essential terms and relations. This ontology can be seen as a domain ontology which allows extension through defined patterns such as generalization, specialization or composition of capabilities.

The high level of formalization of semantic capability models can be used to infer new combinations of CPPS from the explicitly modeled ones that together provide a capability required to produce a product \cite{JHS+_UtilizingSPINRulesto_2018}.

However, there is often a gap between the models of abstract capabilities and executable functions (i.e., \emph{skills}), which are needed to start and control production execution. Often, skills are described separately in a less formal language such as XML.
In \cite{kocher_formal_2020} an integrated capability and skill model in the form of an ontology is presented and the benefits of having a direct connection between capabilities and skills are discussed.

Overall, for a high degree of changeability on control level, approaches based on \emph{capabilities} and \emph{skills} have been proven to be more advantageous compared to classical control approaches \cite{DoWe_EvaluatingSkillBasedControlArchitecture_2019}.

In \cite{ESA+_RoadmaptoSkillBased_2019}, the authors described their initial ideas to use a classical planning approach in a capability-based production in order to find an automation solution for a given task. However, the approach represents more of a roadmap than a concrete solution approach.

%% file: symbolic.tex
\subsection{Symbolic Planning Approaches}
\label{sec:classicalPlanning}
AI planning aims to find a solution to a given problem in form of a sequence of actions leading from an initial to a goal state. Traditional AI planning methods are based on symbolic approaches such as search and logical reasoning. This section focuses on the representation of a planning problem in the Planning Domain Definition Language or as a satisfiability problem which are the basis of most planning algorithms in this field. 

\subsubsection{Planning Approaches based on PDDL} 
The Planning Domain Definition Language (PDDL) was introduced in 1998 for the International Planning Competition (IPC) to express the physics of a domain \cite{aeronautiques1998pddl}. The language follows the declarative, classical paradigm. In PDDL, a planning problem is divided into a domain and a problem description which helps to reuse a domain description for different problems \cite{Anis.92014}. 
Available actions are defined based on the STRIPS-style (Stanford Research Institute Problem Solver \cite{RichardE.Fikes.1971})\cite{aeronautiques1998pddl}, including a set of optionally typed parameters, necessary preconditions and effects. Preconditions and effects are expressed as predicate statements on provided parameters \cite{8972050}. 
As PDDL was initiated to support classical planning, different extensions and versions of PDDL have been developed to increase the expressivity in order to apply PDDL to more realistic problems. With PDDL 2.1 it became possible to consider time using so-called \emph{durative actions} as well as numeric properties. A qualitative model of time was added later with PDDL 3.0 \cite{Anis.92014}. 

In the course of the IPC, different planners were developed to solve problems expressed in PDDL. However, according to the IPC, the participating planners do not need to support all facets of the language \cite{8247683}. These planners are based on different solving strategies, such as search algorithms and logic approaches or a mixture of both.

The problem formulation and especially the formulation of the actions is very intuitive in PDDL and the language is widely used within research of AI planning. However, PDDL is not expressive enough to represent the complexity of real-world CPPS applications. Thus all previous approaches to use existing PDDL-based planners for such problems failed \cite{8247683}.

\smallskip
\subsubsection{Planning Approaches Based on Satisfiability}

The planning approaches introduced in the previous subsection are based on deduction, meaning that a plan is deduced from information given as initial and goal states as well as possible actions that depend on certain preconditions and lead to certain postconditions \cite{kautz_planning_1992}. The problem of \emph{satisfiability} is different in that it aims to find a valid model (i.e., a set of values) that satisfies a given formula. In the case of boolean satisfiability problems (SAT), these formulas are given in propositional logic.
The basis for applying satisfiability for planning problems was laid by Kautz et al. in 1992 \cite{kautz_planning_1992}.

SAT is a well studied problem with many efficient solvers available. However, describing a planning problem using just boolean constraints is difficult. \emph{Satisfiability Modulo Theories} (SMT) is an extension of the boolean SAT problem in which parts of the propositional formula can be expressed using a variety of different so-called theories such as integers, arrays or real numbers \cite{BaTi_SatisfiabilityModuloTheories_2018b}. This makes it much more suited towards industrial applications. A large number of well established solvers for SMT exist (e.g., \cite{de_moura_z3_2008},\cite{cimatti_mathsat5_2013}). These are most often used for applications in software verification \cite{beyer_unifying_2018} and automated theorem proving. While interfacing with these solvers is usually done through APIs, a standardized language for SMT exists with the smt-lib specification \cite{barrett_smt-lib_2011}. Some modern solvers include tools for optimization, this is sometimes referred to as optimization modulo theory (OMT)\cite{LeofanteGAT20}. 

Central to using satisfiability as an approach to planning is how to formulate a problem in a suitable way. An initial formalism for a limited scenario is given in \cite{kautz_planning_1992}, encoding entities as binary variables, actions as predicates and time as integers. The advantage of this approach is that adding additional constraints to the solution---not only to initial and goal state---is rather straightforward. Leofante et al. have a more recent approach using OMT and are focusing on production planning \cite{LeofanteGAT20}. Here the state space is encoded using real valued variables. Possible movements are kept in a transition matrix and the goal is a set of state variables. Together this is a bounded symbolic reachability problem. In order to find an optimal solution it is combined with a reward function specific to the given task. Leofante et al. also provided an implementation to use PDDL to describe a problem and then subsequently solve it using OMT \cite{LeofanteGAT20}. An overview of different modeling strategies for industrial manufacturing using OMT is presented by Roselli et al., however their focus lies on scheduling \cite{roselli_smt_2019}.


%% file: ml.tex
\subsection{Planning Approaches Based on Machine Learning}
\label{sec:machineLearning}
\emph{Machine Learning} (ML) algorithms use data in order to automatically fit a function that solves a given task. In the context of planning, ML can be used to directly learn an optimal plan, to learn useful heuristics for solving a planning problem or to learn functional dependencies between process steps.
 
In the following subsections, the state of the art of ML for planning is presented, categorized by the way the algorithms receive feedback if they perform a task correctly.

\smallskip
\subsubsection{Supervised Learning}

Supervised ML uses domain knowledge in form of labeled datasets to train a model. The weights of the model are adjusted iteratively regarding to prior knowledge. The goal for planning tasks is to find the shortest path from an initial state to a goal state in a specific domain. This chapter only focuses on the most recent approaches, for a broader overview on this topic we refer to \cite{Jimenez.2012}.


The STRIPS-HGN \cite{WilliamShen.} and the ASNet \cite{Toyer.2020} model both rely on a STRIPS description of the planning domain. The STRIPS-HGN learns domain independent heuristics on the delete relaxation of the problem (negative effects are not considered) by encoding the input graph, recursively applying message-passing and decoding the graph representation to extract the heuristic value. 
In training, weights are learned with the Mean-Squared-Error loss against the perfect heuristic h*. 
ASNet transforms a feature representation of the current state into a policy by processing the input through alternating layers of action and proposition modules. 
Due to a domain specific connectivity between the layers, the learned weights of the policy network can be shared for different problems in the domain. ASNet is learned via imitation learning, so that the policy mimics the optimal solution. While these methods work quite well for small problems, they do not output a sequence, but only a heuristic/policy, which serves as input for an off-the-shelf planner.  

Other approaches use (Gated) Recurrent Graph Neural Networks for solving shortest path problems, such as the GGS-NN \cite{Li.17.11.2015} or the Graph2Seq \cite{Xu.03.04.2018}. These models discard the STRIPS representation and are applicable for more general tasks. The GGS-NN model consists of sequentially operating Gated Graph Neural Networks (GG-NN) modules, which each consists of a propagation module to create node representations and an output module. At each timestep, the GG-NN module outputs the node prediction for the current step and the node annotations for the next step until the goal is reached.
The Graph2Seq model maps an input graph to an output sequence. The graph encoder generates a graph embedding, which is decoded sequentially by a Recurrent Neural Network like structure with a special node attention. For the training phase, different graph-path examples are created. In comparison to the former models, these approaches directly learn an output sequence.

So far, all models only work sufficiently for small problem tasks and are not able to solve real world problems.




\smallskip
\subsubsection{Unsupervised Learning} 
Labeling a dataset is a labor-intensive process, which is why real-world datasets in production usually do not contain labels. Thus, there is a broad trend in ML to find methods to exploit knowledge in a dataset in an unsupervised manner. 
As the research community is only beginning to test the potential of unsupervised methods for planning, the approaches are not ready for CPPS yet. However, there are initial works in the domain of generalized planning that have recently addressed the dependence on labels and prior knowledge about the underlying system. 
	
For example, planning instances can be clustered according to their similarity, i.e. whether their solutions share a common structure using an unsupervised method \cite{SJJ17}. The clusters can then be used to assign planning instances to a fitting generalized plan.
	
Rather than using unsupervised ML to learn an optimal plan directly, it can be used to learn general heuristics that simplify the problem and thus help to solve the problem more efficiently. By automatically labeling states on whether they are alive or unsolvable, a mixed integer linear program can be developed \cite{Fra+19}. As the method automatically labels the states by brute force, it only works on problems with a small number of reachable states. 
Building on that work, a qualitative numerical planning problem can be learned, which can then be solved using a SAT planner more efficiently \cite{FBG21}. 
	
A key challenge is to detect states that are unsolvable. By automatically labeling states on a few small instances of the planning domain with an exhaustive exploration, a heuristic can be learned that helps detecting similar states in more complex planning domains \cite{SFS21}. 


\smallskip
\subsubsection{Reinforcement Learning}

In Reinforcement Learning (RL), the goal is for an agent to learn the optimal policy to maximize the rewards it receives \cite{Kuhnle.2019}. So far, complex, real-world applications for RL have rarely been considered, and the tractability of an RL agent's behavior has not yet been studied \cite{Kuhnle.2021}. Nevertheless, RL offers a potential to solve complex and dynamic decision problems as an alternative to mathematical approaches \cite{Kuhnle.2021}.

In particular, (i) applications with a limited scope (in terms of the number of states and actions), (ii) responsive real-time decision systems, (iii) complex environments that can hardly be described in detail, and (iv) abundant or easily generated training data, are useful features \cite{Kuhnle.2019}.

In order to increase robustness, Kuhnle et al. use a Trust Region Policy Optimization (TRPO) agent to study two production problems in a system of eight machines: order sequencing and route planning \cite{Kuhnle.2021}. By giving dense reward after each complete iteration of state, action and reward, good results in detecting valid and invalid actions can be achieved through a lower training cost.

A different approach is taken by Eysenbach et al., who use the so-called \emph{Search on the Replay Buffer} (SoRB) method to learn a goal-conditioned policy and afterwards apply graph search to plan high-dimensional tasks over longer time horizons. The approach was tested using a complex visual navigation task. Using automatic wayfinding by observations in the replay buffer and subsequent graph search, a reasonable sequence of waypoints can then be planned.\cite{Eysenbach.2019} 

Rivlin et al. show that planning strategies involving graph neural networks can be learned that are generalizable, without considering heuristics or existing solutions. By creating a state-goal graph, the learning process is improved and general applicability in longer instances is enabled.\cite{Rivlin.2020}


%% file: scheduling.tex
\subsection{Scheduling} \label{sec:scheduling}

Scheduling is a well-established field in AI and production respectively, a survey of the applied algorithms is given by \cite{zhang2019review}. 
Several publications cover shop-floor applications, for example \cite{akbar2020metaheuristics}.
Typical constraints in the field of production have also been modeled. Different algorithms have been evaluated (e.g.,\cite{ELMARAGHY2000186,Sierra2015NewSG}). Existing approaches can also differentiate between automated and non-automated processes \cite{doi:10.1080/00207543.2013.831220}. Also different optimization criteria have been evaluated \cite{LI2016113}.

%% file: discussion.tex
\section{Discussion}
\label{sec:discussion}

In this section, the suitability of existing AI-algorithms for CPPS planning is evaluated. For this purpose, for each of the AI solution steps from Section \ref{sec:introduction}, the suitability of the existing AI algorithms from Section \ref{sec:ai} is evaluated with regards to the requirements presented in Section \ref{sec:requirements}. Table \ref{tab:planning-approaches} shows an overview of this evaluation.

\begin{table*}
    \centering
    \caption{Comparison of the requirements with existing research around the three main steps of an AI planning solution (see Fig. \ref{fig:planning-steps}).
    \CIRCLE/\LEFTcircle/\Circle: Fully/Partially/Not covered  \quad \quad - : Not applicable }
    \label{tab:planning-approaches}
    \renewcommand*{\arraystretch}{1.45}
    \begin{tabularx}{0.85\textwidth}{X c c c c c c}
    \toprule
        &
        \thead{R1 \\ (Optimization)} & 
        \thead{R2 \\ (Dependencies)} &
        \thead{R3 \\ (Loops)} &
        \thead{R4 \\ (Explainability)} &
        \thead{R5 \\ (Low Effort)} &
        \thead{R6 \\ (Data)} \\
        \midrule
        \textbf{Step 1: Input Models} & \LEFTcircle & \LEFTcircle  & - & \CIRCLE & \LEFTcircle & \Circle 
        \\
        \textbf{Step 2: Planning} &  &  &  &  &  & 
        \\ 
        \textbf{\qquad Symbolic Planning} & \CIRCLE & \LEFTcircle & \LEFTcircle & \LEFTcircle & \LEFTcircle & \Circle 
        \\
        \textbf{\qquad Planning based on ML} & \LEFTcircle & \CIRCLE & \LEFTcircle & \Circle & \Circle & \CIRCLE 
        \\
        \textbf{Step 3: Scheduling} & \CIRCLE & - & - & \CIRCLE & \CIRCLE & \CIRCLE 
        \\
        \bottomrule
    \end{tabularx}
\end{table*}

\subsection{Step 1: Input Models}
In the following, we deal with input models for planning (see Step 1 in Fig. \ref{fig:planning-steps}) and evaluate whether existing approaches to model CPPS semantically as presented in Section \ref{sec:models} are able to meet the requirements defined in Section \ref{sec:requirements} and can be used to automate a planning solution.

In order to optimize for certain criteria such as total cost or lead time, relevant information about each machine and each process has to be captured within the input models. While statements about individual costs or times are comparatively easy to express, capturing interrelationships, for example between conflicting variables, is difficult. Furthermore, there is no standardized approach to use semantic models in the subsequent steps of planning and scheduling. Thus, R1 can be regarded as partially fulfilled.

Semantic capability models aim to capture e.g. processes and the corresponding dependencies between inputs and outputs in a formal way. As with optimization criteria, modeling mathematical relationships in such models is complicated, which is why R2 is also only partially fulfilled.

Some of the approaches to semantic capability modeling presented in Subsection \ref{sec:models} support a decomposition of higher-order capabilities into a sequence of more detailed ones. Recurring invocations of capabilities may be modeled in this way. However, dynamically finding recurring invocations or loops needs to be done by a planning algorithm, so R3 is not really applicable to models.

Highly formal models of machines and their capabilities as well as products offer high potential as a basis for explainable planning systems. Especially when considering ML approaches---which are often regarded as \emph{black boxes}---ontologies are seen as a promising means to offer a symbolic justification of decisions \cite{SoLe_AligningArtificialNeuralNetworks_2021} (R4).

Creating extensive semantic models of machines and their capabilities is a tedious and error-prone task. At the same time, there are only few experts who are able to do this. Initial approaches like \cite{KHC+_AutomatingtheDevelopmentof_2020} or \cite{KJF_AMethodtoAutomatically_2021} make use of existing engineering artifacts like 3D models of machines or PLC code to automate some of the tasks of creating a semantic capability model. R5 is therefore considered partially fulfilled. 

Learning semantic models that can be used by production planning algorithms is in its infancy with \cite{HGR+_OntologyBasedSkillDescriptionLearning_2020} presenting an initial approach to capability description learning via
inductive logic programming. In terms of Step 1, R6 is still an open challenge.


\subsection{Step 2.1: Symbolic Planning}

In the following, we evaluate whether existing AI algorithms from Section \ref{sec:classicalPlanning} can be used to improve---or automatize---the planning step (see Step 2 in Fig. \ref{fig:planning-steps}).
The first requirement can be met by many planners based on PDDL as well as planners based on satisfiability. In case of a SMT-solver, for example variables for costs or quality can be easily modeled and taken as optimization criteria---given that the problems are of linear nature as most planners do not support OMT with nonlinear functions (R1). While there are some extension to planning algorithms with regard to continuous inputs/outputs, they are still mainly used for purely symbolic representations. Especially process parameters are hard to integrate. But from a theoretical point of view, as determining valid variable assignments is an inherent feature of approaches based on satisfiability, those solvers should be able to identify a valid set of process parameters, as far as modeling allows (R2). The third requirement is neither fulfilled by PDDL-solvers nor by solvers based on satisfiability. While they can allow for recurrent calls to functions, recognizing loops still poses a problem (R3). The explainability (R4) of the planners depends on the underlying algorithm. Most symbolic planners use algorithms whose choices at each decision point are deterministic and repeatable \cite{MDD_ExplainablePlanning_2017}. Creating symbolic planning models is a tedious task. But underlying algorithms do not need any user interaction. And as languages such as PDDL and smt-lib are standardized, solvers can be changed without adaptions to the model (R5). However, as all symbolic methods, they are not well suited to integrate a plethora of data points since they rely on a defined ground truth (R6). 
 

\subsection{Step 2.2: Planning based on ML}

As described in Section \ref{sec:machineLearning}, ML offers a variety of approaches to automatize the planning step. However, most methods still require a relative small and manageable system, thus modeling all necessary machine capabilities of a complex CPPS is not possible yet. Although there are differences between supervised, unsupervised and reinforcement learning, we analyse them together in this high-level overview.
Given sufficient data, a ML model can learn how a specific planning configuration impacts characteristics such as resource consumption or the price. In order to be used for optimization (R1), these cost functions must be extrapolatable, i.e. they must predict cost for operation points not covered by the data. This is a challenge for most ML algorithms \cite{Bishop.2006}.
The second requirement can be met, as ML algorithms can detect complex dependencies between inputs, outputs and process parameters, even if these functional dependencies are not obvious (R2). 
In general, the algorithm benefits from the use of prior knowledge of the system to detect these dependencies, which is the focus of supervised learning.
While an ML algorithm might learn to perform a functionality multiple times, learning when and how often to perform these loops in a control flow is still challenging (R3).
ML algorithms can learn complex functional dependencies of a system on their own. The overall behavior of the resulting model is generally hard to assess, thus ML algorithms often suffer from a lack of explainability (R4).
Deploying ML algorithms can require lots of implementation effort, as choosing and configuring algorithms can be challenging. 
There are hardly any ready-to-use solutions and the level of standardization is lower than for symbolic procedures (R5). All methods require an automation engineer to implement some knowledge about the specific CPPS, even if some approaches like unsupervised learning try to limit this effort.
Finally, ML algorithms utilize existing data and are thus suited to learn models reflecting the underlying CPPS (R6). 

\subsection{Step 3: Scheduling}

In the following, we evaluate whether existing AI-algorithms from Section \ref{sec:ai} can be used to improve---or automate---the scheduling step (Step 3 in Fig. \ref{fig:planning-steps}).
As the goal of scheduling is to optimize the production under certain criteria such as minimizing the makespan, scheduling algorithms are able to support optimization (R1). 
Since during the scheduling step no continuous interdependencies are used anymore, scheduling algorithms do not need to fulfill the second requirement. The same applies to the third one. The decisions as well as the results of scheduling algorithms are repeatable and traceable (R4). Just as with planning, the algorithms are not usually changed during scheduling, thus they do not need support by a user. Scheduling is based on the input model which has to be adjusted instead (R5). The scheduling algorithms can benefit from certain data provided by the MES such as machine occupancy (R6). An example of a successful integration of a scheduling solution with the MES which considers real-time production information has been presented by \cite{Zhou2018AnED}.  


\section{Research Agenda}
From the analysis of the requirements given in Section \ref{sec:requirements} against the steps of Fig. \ref{fig:planning-steps}, a research agenda with five different fields of future research (FR) is derived:

\smallskip

\noindent \emph{FR 1: Ontologies and Functional Dependencies} 

One way to represent knowledge about CPPS in a machine-interpretable way is the use of ontologies, which are also interpretable by humans. However, some functional dependencies between continuous values can hardly be captured using ontologies. On the other hand, ML algorithms are good at learning these functional dependencies from data. As both methods are in large parts complementary, which can also be seen in Table \ref{tab:planning-approaches}, we miss an integration between semantic models and ML.



\smallskip
\noindent \emph{FR 2: Planning Approaches for CPPS} 

Current planning approaches are not designed to make use of CPPS models. Even without the integration between symbolic and subsymbolic approaches (see FR 1), CPPS models can already represent much more information than can be processed by current planning approaches. Both languages / theories such as PDDL or SMT and their corresponding algorithms need to be improved to prevent a loss of information and guarantee better planning results.


\smallskip
\noindent \emph{FR 3: Explainability} 

The acceptance of AI solutions depends on the explainability of the results. So far, only partial solutions for specific algorithms exist. Explainability of results on all steps of a planning solution is needed. It can be improved based on semantic input models of systems. Thus, future research should aim to incorporate semantic information from the CPPS models into subsequent planning algorithms and their results.

\smallskip
\noindent \emph{FR 4: Detecting Loops} 

Many processes in industrial production involve recurring process steps, e.g., re-working of sub-optimal products. While multiple occurrences of a process step can be identified by many planning approaches, explicitly detecting loops and their conditions is only handled by some initial algorithms. Explicit detection of loops could reduce costly replanning, which is particularly relevant as batch sizes in industrial production decrease.

\smallskip
\noindent \emph{R5: Real Benchmarks} 

The field of AI planning has brought a high number of approaches which were never tested on real industrial applications. An end to end case study covering all three steps of modeling, planning and scheduling would shed more light on the maturity of the discussed approaches. For true comparability, a corresponding data set is needed, ideally provided as a benchmark to the research community.

%% file: conclusion.tex
\section{Conclusion}
\label{sec:Summary}
Regarding our research question, it can be noted that semantic models of CPPS can currently not be consumed by planning and scheduling algorithms. Even though there are obvious benefits of such an integration (e.g., automated problem generation, possible explainability), there currently is no interrelation between the two fields.
In this contribution, we established a common notion to production planning which comprises both semantic CPPS models as well as the actual planning/scheduling algorithms. Requirements for future CPPS planning solutions were introduced and existing research in this area was discussed with respect to these requirements. The approaches covered can be categorized into two main categories: While the first one contains approaches of symbolic AI such as semantic models, \emph{PDDL} and \emph{SMT}, the second category consists of sub-symbolic AI approaches from the field of ML.

While we tried to give a concise summary of the approaches and their compliance with regard to the requirements (see Tab. \ref{tab:planning-approaches}), there are limitations when it comes to discussing a whole category such as \emph{PDDL} as there exists a variety of individual approaches using PDDL which might fulfill (or neglect) specific requirements. Therefore, Tab. \ref{tab:planning-approaches} can only give a coarse overview of a category which might not reflect the particularities of every individual approach of that category.

Overall, a strong integration of (semantic) CPPS models with subsequent algorithms for planning and scheduling is missing. Most, if not all, ML-based planning approaches are in their infancy and are being tested in simplified scenarios that cannot be compared to realistic planning problems. Symbolic methods have a higher degree of maturity due to their longer research history but still do not fulfill all requirements of a flexible production based on CPPS. 
A research agenda has been derived which covers relevant fields of research that need to be tackled in order to overcome these limitations.  In our future work, we will address items on this agenda.